\documentclass[a4paper,11pt]{article}
\usepackage[margin=1in]{geometry}
\usepackage{amsmath}  
\usepackage{url}
\pdfoutput=1

\usepackage{microtype}
\usepackage{graphicx}
\usepackage{subfigure}
\usepackage{booktabs} 

\usepackage{framed}
\usepackage{xcolor}
\usepackage{tcolorbox}

\usepackage{tablefootnote}

\usepackage{amsmath,amsthm,amssymb,amsfonts}
\usepackage{bm}
\usepackage{pifont}
\usepackage{stackengine}

\usepackage{algorithm}
\usepackage{algorithmic}
\usepackage{enumerate}

\theoremstyle{plain}
\newtheorem{theorem}{Theorem}[]
\newtheorem{lemma}[]{Lemma}
\newtheorem{corollary}{Corollary}[]
\newtheorem{definition}{Definition}[]
\newtheorem{assumption}{Assumption}[]

\def\x{{\textbf x}}
\def\z{{\textbf z}}

\def\g{{\textbf g}}

\def\w{{\textbf w}}
\def\v{{\textbf v}}

\def\R{{\mathbb{R}}}

\def\O{{\mathcal O}}

\title{Fed-QSSL: A Framework for Personalized Federated Learning under Bitwidth and Data Heterogeneity}
\author{
    Yiyue Chen\textsuperscript{\rm 1},
    Haris Vikalo\textsuperscript{\rm 1},
    Chianing Wang\textsuperscript{\rm 2}
    \thanks{This work was supported by Toyota Motor North America.}
}
\date{
    \textsuperscript{\rm 1}The University of Texas at Austin\\
    \textsuperscript{\rm 2}Toyota InfoTech Lab USA\\
    yiyuechen@utexas.edu,  hvikalo@utexas.edu,
    johnny.wang@toyota.com
}
\begin{document}
	\maketitle
		\vskip 0.3in
	
\begin{abstract}
	Motivated by high resource costs of centralized machine learning schemes as well as data privacy concerns, federated learning (FL) emerged as an efficient alternative that relies on aggregating locally trained models rather than collecting clients' potentially private data. In practice, available resources and data distributions vary from one client to another, creating an inherent system heterogeneity that leads to deterioration of the performance of conventional FL algorithms. In this work, we present a federated quantization-based self-supervised learning scheme (Fed-QSSL) designed to address heterogeneity in FL systems. At clients' side, to tackle data heterogeneity we leverage distributed self-supervised learning while utilizing low-bit quantization to satisfy constraints imposed by local infrastructure and limited communication resources. At server's side, Fed-QSSL deploys de-quantization, weighted aggregation and re-quantization, ultimately creating models personalized to both data distribution as well as specific infrastructure of each client's device. We validated the proposed algorithm on real world datasets, demonstrating its efficacy, and theoretically analyzed impact of low-bit training on the convergence and robustness of the learned models.
\end{abstract}
\section{Introduction}
Federated learning (FL) \cite{mcmahan2017communication, kairouz2021advances} recently emerged in response to the challenges in data privacy, storage cost and computation commonly faced by the conventional (centralized) learning systems. In centralized learning, a server collects and stores data to be used for model training, raising concerns regarding potential leakage of sensitive information and inefficient utilization of the storage and computation resources. Lately, advancements in software and hardware technologies have equipped smart edge devices with computational power that enables local data processing, allowing implementation of distributed frameworks such as FL in a variety of real-world applications \cite{chen2019communication}. In FL systems, participating devices collaboratively learn a model while preserving privacy by training on local data that remains private. A server aggregates local updates to obtain a global model which is then distributed to the clients; in turn, the clients continue local training using the latest global model as a starting point. In addition to the conventional supervised learning, federated learning is suitable for meta-learning and unsupervised learning problems \cite{jiang2019improving, servetnyk2020unsupervised}. 

Heterogeneity in local data distribution, as well as imbalance in the computation and memory capabilities of the participating devices, present major challenges to federated learning \cite{zhao2018federated, yoon2022bitwidth}. Practical scenarios where both sources of heterogeneity occur include healthcare disorder prediction for patients with different profiles and monitoring devices, transportation systems over different vehicles and terrain, and indoor/outdoor air quality detection, to name just a few. Several recent FL techniques aim to 
alleviate the impact of data, model or device heterogeneity \cite{jiang2020federated, lin2020ensemble, diao2020heterofl, wang2022does}. However, variations in bitwidth available to the clients that participate in training have been much
less studied \cite{yoon2022bitwidth}. To our knowledge, there exists no prior work in literature investigating FL in 
setting where both the local data and device bitwidth are heterogeneous -- the focus of our work.


On one hand, clients' devices collect and/or generate data at different times and locations, leading to discrepancy in data amounts and distribution. When distributed devices train locally, each learned model optimizes an objective specified in regard to a local dataset. If the data across participating devices is non-IID, the devices end up optimizing distinct objectives which generally adversely affects 
training convergence and, ultimately, negatively impacts performance of the resulting global model trained by classical FL algorithms such as FedAvg \cite{mcmahan2017communication, li2019convergence}. This is exacerbated when a client's dataset contains only a small subset of classes or few data points, leading to an insufficiently expressive local model. In turn, using such models in the aggregation step at the server leads to a global model that lacks robustness and in general may significantly underperform its centrally trained counterparts \cite{zhao2018federated}.

On the other hand, to satisfy the constraints on local compute and memory footprint as well as on the communication bandwidth between clients and the central server, each client needs to train a local model in low bitwidth operations and store the updated model in low bitwidth \cite{yoon2022bitwidth}. When the clients' devices have different bitwidth capabilities, a number of novel challenges arise including: (1) Quantized training conducted at lower precision levels does not necessarily lead to expressive local model; (2) the server needs to aggregate local models that are represented by different number of bits; and (3) following the aggregation of local updates, the server should communicate to clients the new global model re-quantized at levels matching the capabilities of different devices. Prior works that deploy model compression in distributed settings focus on reducing the communication cost \cite{reisizadeh2020fedpaq, chen2021communication, chen2021decentralized}; those methods utilize full precision weights in the training process rather than trying to train under weight precision constraint. To address the problem of aggregating models trained at different precision, \cite{yoon2022bitwidth} proposed a progressive weight de-quantizer that transforms low precision models to their full precision versions before aggregating them. The de-quantizer there requires the server to train DenseNet for de-quantizing low precision models; the method assumes the local data distribution is the same across different clients and thus does not apply to the heterogeneous data settings studied in this paper.
Note that all of these methods are restricted to supervised learning.

The contributions of this paper are summarized as follows:

\begin{enumerate}

\item For the first time, a novel FL problem characterized by heterogeneous local data distributions and varied capabilities of clients' devices is studied.
This is the first work to consider the non-trivial task of aggregating models which due to varied bitwidth and heterogeneous data have incompatible parameters and different expressiveness.
While prior work addresses either source of heterogeneity in isolation of the other, to our knowledge the combination of the two has not been investigated previously. Indeed, new challenges set forth by the combination of the heterogeneity sources render the prior techniques ineffective.

\item We present a new FL framework, Fed-QSSL, that enables learning personalized and robust low-bitwidth models in settings characterized by diverse infrastructure and data distributions. Fed-QSSL combines low-bitwidth quantization training with self-supervised 
learning at the client side, and deploys de-quantization, weighted aggregation and re-quantization at the 
server side.

\item We theoretically analyze the impact of low-bit training on the convergence and robustness of learned
models. In particular, we present a bound on the variance of the quantization errors in local and global training and investigate the associated convergence speed. The analysis demonstrates that low-bit training with 
Fed-QSSL on heterogeneous data allows learning meaningful representations. 

\item Finally, the efficacy of the proposed algorithm is tested in a variety of experiments.

\end{enumerate}

\subsection{Related work}

{\bf Bitwidth heterogeneity in FL.} To address local infrastructure constraints, \cite{diao2020heterofl} rely
on models with simple architectures and low-bit parameter representation. However, existing prior work
typically assumes that clients train models of same precision, e.g., the same bitwidth, which may often
be violated in practice since the participating devices generally have different computational power and/or memory 
footprint. The varying infrastructure capabilities of the clients' devices imply that the models they are able to 
deploy differ in size, i.e., devices with more computational power and memory allow models of higher 
precision (e.g., 16-bit and 32-bit) while less capable devices may only handle low-precision models (e.g.,
those storing parameters in 2-bit or 4-bit precision). To satisfy local infrastructure constraints, not only should 
a model be represented and stored at reduced precision levels but the training process should also rely on 
operations at such levels (i.e., perform quantized training in low-bitwidth). Recently, \cite{yoon2022bitwidth} 
studied the bitwidth heterogeneity problem in FL and proposed a progressive de-quantization of the received
local models prior to their aggregation into a global model.

{\bf Quantized neural network training.} 
When the compute/storage resources are limited, on-device training of deep neural networks at full precision is rendered infeasible. To this end, recent works have explored model size reduction and the training that leads to low bitwidth weights/activations. These include binarized neural networks (BNN) and XNOR-Net which binarize the weights and activations of convolutional layers to reduce the computation complexity in the forward pass \cite{hubara2017quantized, rastegari2016xnor}.
Specifically, in the forward pass the computationally expensive convolutional operations are implemented via bitwise operation kernels that evaluate the dot product of binary vectors $\x$ and $\mathbf{y}$
\begin{align*}
    \x \cdot \mathbf{y} = \mathrm{bitcount}(\mathrm{and} (\x, \mathbf{y})), x_i, y_i \in \{0, 1 \} \ \forall i,
\end{align*}
where $\mathrm{bitcount}(\cdot)$ counts the number of bits in its argument. Such a kernel can be generalized to accommodate operations on $M$-bit fixed-point integer sequence $x$ and $K$-bit fixed-point integer sequence $y$, incurring computation complexity of $\O(MK)$ (i.e., the complexity is proportional to the vector bitwidths). Related work \cite{zhou2016dorefa} presents DoReFa-Net which reduces the backpropagation computation cost by quantizing gradients in the backward pass, and studies the effect of the weight, activation and gradient bitwidth choice on the model performance. Subsequent works include techniques that deploy training without floating-point operations, gradient clipping, and training under mixed precision \cite{wen2017terngrad, zhang2017zipml, zhu2020towards, zhang2020fixed}. While all of the aforementioned methods investigate quantization in supervised learning settings, recently there has been interest in quantizing models for self-supervised learning. In \cite{cao2022synergistic}, the authors propose self-supervised and quantization learning (SSQL), which contrasts features learned from the quantized and full precision models in order to provide reasonable accuracy of quantized models and boost the accuracy of the full precision model. While this work can balance the accuracy and bitwidth in resource constrained settings, it still requires full precision operations and gradient storage during training. 


{\bf Non-IID data and distributed self-supervised learning.} There have been multiple efforts to ameliorate the impact of data heterogeneity on the performance of distributed learning systems \cite{karimireddy2020scaffold, ghosh2020efficient, li2022federated}. Recently, self-supervised learning (SSL) has been shown effective in distributed settings where the data is large-scale and imbalanced \cite{wang2022does}. In contrast to supervised learning, which requires large amount of labeled data for model training, self-supervised learning defines a pre-train task that enables extracting expressive representations; those representations can then be utilized in various downstream tasks in computer vision and natural language processing \cite{chen2020simple,chen2021exploring}. While the majority of self-supervised learning methods focus on the centralized setting where the data is collected for central training, recent works have attempted to bridge the self-supervised and federated learning \cite{he2021ssfl, zhuang2022divergence, makhija2022federated}.  

\section{Preliminaries}
Let us consider a federated learning system with $n$ clients, where the clients rely on privately owned data to collaboratively learn a global model. Rather than communicating the data, the clients send to a coordinating central server the models trained locally. In turn, the server forms a global model by aggregating the received local models, and re-distributes the global model back to the clients for further local training. Let $D_k = \{x_{k, i} \}_{i=1}^{|D_k|}$ denote the data owned by client $k$, where $x_{k, i} \sim \mathcal{D}_k$ is a $d$-dimensional data point in $D_k$ and $|D_k|$ denotes the number of local data points. The distribution $\mathcal{D}_k$ of data varies from one client to another, leading to system-wide data heterogeneity. The union $D = \bigcup_{k=1}^n D_k$ is assumed to be a dataset of uniform distribution, i.e., the data classes in $D$ are balanced. 

In self-supervised learning problems, an embedding function $f_{\w}(\cdot)$, parameterized by $\w$, is learned as a feature extractor for extracting expressive representations from data. In particular, we will denote the representation vector for point $x$ by $f_{\w}(x)$; such a representation vector is then used for downstream tasks such as image classification or object detection. Contrastive learning is a popular methodology for identifying the embedding function $f_{\w}(\cdot)$ \cite{chen2020simple}. The aim of contrastive learning is to find an embedding function such that the distance between the learned representations of data point $x$ and its positive signal $x^+$ (generated from $x$) is small, while the distance between the representations of $x$ and negative samples $x^{-}$ (extracted from the training batch) is large.\footnote{Given a data sample $x$, its positive signal may be a variant obtained by adding noise, changing color, or clipping; negative signals include data samples from different classes, or simply other data samples in the same training batch.}
A commonly used objective in contrastive learning is the InfoNCE loss; specifically, the local objective for client $k$ is
\[
\mathcal{L}_{CL, k}(\w) = \frac{1}{|D_k|}\sum_{i=1}^{|D_k|} -\log ( \frac{\mathbb{D}_{k, i}^+}{ \mathbb{D}_{k, i}^+ + \sum_j \mathbb{D}_{k, i, j}^-}), 
\]
where
$\mathbb{D}_{k, i}^+ = \exp(-\mathbb{D}(f_{\w}(x_{k, i}), f_{\w}(x_{k, i}^+))/\tau$, $\mathbb{D}_{k, i, j}^- = \exp(-\mathbb{D}(f_{\w}(x_{k, i}), f_{\w}(x_{k, j}^-))/\tau$, and
$\tau >0$ is a temperature parameter controlling the sharpness of the exponential term. The function $\mathbb{D}(\cdot)$ is typically defined as the cosine distance, i.e., $\mathbb{D}(z_1, z_2) = -\frac{z_1 \cdot z_2}{\|z_1 \| \|z_2 \|}$. Note that instead of the negative sample term, an alternative SSL strategy adds to the objective a feature prediction function $g(\cdot)$ that helps avoid potential collapse i.e. low accuracy of the trained classifier layer \cite{chen2021exploring}; in that case, the objective becomes 
\[
\mathcal{L}_{siam, k}(\w) =  \frac{1}{|D_k|}\sum_{i=1}^{|D_k|} \mathbb{D}(g(f_{\w}(x_{k, i})), f_{\w}(x_{k, i}^+)).
\]

In federated learning, each client has a local data source $D_k$. The clients collaboratively search for $\w$ that minimizes the global objective function
\begin{align}
    \mathcal{L}(\w) = \sum_{k=1}^n \frac{|D_k|}{|D|}\mathcal{L}_k(\w).
\end{align}
To solve this distributed optimization without data sharing, numerous federated learning schemes have been proposed including the FedAvg algorithm \cite{mcmahan2017communication}. At each iteration of FedAvg, central server samples a subset of clients which then run $E$ local update epochs. The server collects updated models from the selected clients and aggregates them to form a global model, which is distributed to all clients for further local training/inference. The number of local training epochs, $E$, controls the computation-communication tradeoff, i.e., larger $E$ leads to less frequent communication. 

We further assume that participating clients have devices with varied compute, memory and communication capabilities, and therefore train and deploy models at different bitwidth precision. Let $s_k$ denote the number of bits used to represent parameters $\w_k$ of the model deployed by client $k$. When training, client $k$ utilizes bit convolutional kernels and conducts operations on $s_k$-bit objects rather than performing full precision (i.e., $32$-bit floating-point) operations. When aggregating the collected local models into global model $\w_G$, the server takes into consideration that the models are trained at different precision levels. The server itself is assumed to be capable of performing operations at full precision. Finally, the global model is compressed to various low bitwidth representations and sent to the corresponding clients, i.e., client $k$ receives updated model in $s_k$ bit representation.

In our proposed framework, client $k$ performs low-bit training to learn parameters $\w_k$ of its local model; note that the low-bit training and quantization are deployed in all steps, including the forward pass, backpropagation and model update. In each round of training, model parameters in $s_k$-bit representation are fed into the forward pass via bit convolution kernels, and the gradient computed from the backward propagation is quantized and stored. After updating the model by using gradient descent, the model parameters are quantized to $s_k$ bits and stored. 

\section{Algorithm}
The proposed federated quantized self-supervised learning scheme consists of training procedures deployed at clients' devices and the server, as detailed next.

At each client's side, a feature extraction model is trained in low bitwidth with layer-wise quantization in both forward and backward propagation pass. We denote the model parameters of client $k$ in training epoch $t$ by $\w^{(Q)}_{k, t}$; each parameter is represented using  $s_k$ bits. The codebook used for quantizing parameters of each layer of the model, $\w^{(Q)}_{k, t}$, has cardinality $2^{s_k}$. The quantization centers (i.e., codebook entries) are typically determined via uniform quantization \cite{zhou2016dorefa},
\begin{align*}
     r_o = \mathrm{quantize}_s(r_i) =  \frac{1}{2^s-1}\mathrm{round}((2^s-1)r_i),
\end{align*}
which uses a uniform quantizer $\mathrm{quantize}_s$ to map a real-valued input $r_i \in [0, 1]$ to an $s$-bit output $r_o \in [0, 1]$. In general, parameters of deep neural networks do not follow uniform distribution and may vary in range. Therefore, the parameters may first need to be non-linearly transformed, e.g., by applying $\tanh$ function to limit their range to $[-1, 1]$. If $\tanh$ function is used, the quantized model weights are found as
\begin{align*}
     r_o = 2\mathrm{quantize}_s(\frac{\tanh{r_i}}{2\max(\tanh{r_i})}+\frac{1}{2}) - 1, 
\end{align*}
where the maximum is taken over all parameters in the same layer. In the forward pass, we apply normalization to ensure that $\frac{\tanh{r_i}}{2\max(\tanh{r_i})}+\frac{1}{2} \in [0, 1] $, where $\max(\tanh{r_i})$ represents the maximum taken over all parameters in this layer. Following the quantization operation, output $r_o$ is in $[-1, 1]$, which improves the stability of training.

In addition to quantizing model parameters, the proposed scheme quantizes activations as well. In particular, after the output of a layer passes through an activation function that guarantees the output to be within $[0, 1]$ range, the activation is quantized using $\mathrm{quantize}_s$. 

To quantize gradients in backward propagation, we use layer-wise quantile quantization method. In particular, we rely on the inverse of the cumulative distribution function $F_{\g}$ of the gradients in a layer to define the quantile function $Quantile_{\g} = F^{-1}_{\g}$. The quantization centers are then specified by the end points of each histogram bin, $Quantile_{\g}(\frac{i}{2^s-1})$, for $i = 0, \dots, 2^s-1$. It has been observed in prior work on low-bitwidth training \cite{zhou2016dorefa} that to avoid significant performance degradation, the number of bits for gradient quantization should be no less than the number of bits for weight quantization (and may in fact need to be greater). In our algorithm, local gradient quantization uses $2$ bits more than local weight quantization, e.g., if client $k$ uses $s_k$ bits for weight/activation quantization then it uses $s_k+2$ bits for gradient quantization. Following the backward pass, the computed estimate of stochastic gradient, $\g_{k, t}$, is immediately quantized to $\g^{(Q)}_{k, t}$. Finally, updated weights are quantized according to $\w^{(Q)}_{k, t+1} = Q(\w^{(Q)}_{k, t} - \alpha_t \g^{(Q)}_{k, t})$.

\begin{algorithm}[tb]
\caption{Fed-QSSL: Federated quantized self-supervised learning under bitwidth and data heterogeneity}
\label{alg:algorithm2}
\textbf{Initialization}: Clients bitwidth configuration $\{s_k \}_{k=1}^n$, local dataset $\{ D_k\}_{k=1}^n$, initial models $\{\w^{(Q)}_{k, 0} \}_{k=1}^n $, global data buffer $D_{g}$, $t = 0$ \\
\textbf{Parameter}: number of local training epoch $E$, step size $\{\alpha_t\}$
\begin{algorithmic}[1] 
\FOR{each round $r = 1, 2, $}
\STATE $\{$Local training at clients $\}$
\FOR{each client $k$ in parallel}
\FOR{$t = 0, 1, \cdots, E-1$}
\STATE Compute the compressed gradient $\g^{(Q)}_{k, t+(r-1)E}$
\STATE Update the low-bit weights $\w^{(Q)}_{k, t+1+(r-1)E} = Q(\w^{(Q)}_{k, t+(r-1)E} - \alpha_{t+(r-1)E} \g^{(Q)}_{k, t+(r-1)E}) $
\STATE $t = t+ 1$
\ENDFOR
\ENDFOR
\STATE $\{$Global operations at the server $\}$
\STATE The server collects local models $\{\w^{(Q)}_{k, rE} \}_{k=1}^n$
\STATE The server de-quantizes $\{\w^{(Q)}_{k, rE} \}$ using $D_{g}$ to obtain $\{\w'^{(Q)}_{k, rE} \}$ and loss $\{L_{DQ_k} \}$ for all $k \in [n]$
\STATE The server aggregates  $\w_{rE, G} = \sum_{k=1}^n p_k \w'^{(Q)}_{k, rE}$ using DQ-based losses
\STATE The server re-quantizes the aggregation using $D_{g}$ to $s_k$-bit model and distributes to client $k$ for all $k \in [n]$
\ENDFOR
\STATE \textbf{Post-training} Each client freezes the trained feature extractor and trains local linear probe (i.e., classification layer) for downstream tasks
\end{algorithmic}
\end{algorithm}

In each communication round the server collects and de-quantizes local models, aggregates the de-quantized models, re-quantizes the resulting model to low bitwidth and finally distributes the models to clients. To elaborate, after the central server collects local low-bitwidth models, it uses a small buffer of unlabeled data $D_g$ to de-quantize the collected local models to full precision. The de-quantization process differs from one local model to another and can be performed in parallel. In the scenario without bitwidth constraints, de-quantization can be viewed as a fine-tuning step on uniform unlabeled data; simulation results presented in the next sections demonstrate that such fine-tuning improves robustness in the full-precision model scenario. 

After de-quantizing the collected models, the server forms their weighted average, $\w_{t, G} = \sum_{k=1}^n p_k \w'^{(Q)}_{k, t}$. The weights reflect performance discrepancy among local models caused by both local bitwidth and data heterogeneity, and are computed as 
\begin{align*}
    p_k = \frac{e^{-\mathcal{L}_{DQ_k}}}{\sum_{j=1}^n e^{-\mathcal{L}_{DQ_j}}}
\end{align*}
where $\mathcal{L}_{DQ_k}$ represents the last epoch loss in the de-quantization process of model $k$. 
The server re-quantizes global model $\w_{t, G}$ to an $s_k$-bit version denoted by $\w^{(Q)}_{k, t}$ while running fine-tuning epochs on a uniformly distributed dataset stored in the global buffer $D_g$. The fine-tuning epochs use low-bit operations in order to guarantee that the fine-tuned model stays in $s_k$-bit representation. Note: both de-quantization and re-quantization use $D_g$ for fine-tuning but de-quantization runs full precision operations while re-quantization runs low-bit operations.

After completing pre-training and learning the feature extractor model, each client freezes parameters of the feature extractor model and trains a low-bitwidth classifier on its local (labeled) data. 

The described procedure, including both the steps executed by clients as well as those executed by the server, is formalized as Algorithm \ref{alg:algorithm2}.

{\bf Remark.}
While Algorithm \ref{alg:algorithm2} assumes full client participation at each communication round, it is straightforward to incorporate random client sampling as used by the vanilla FedAvg algorithm in settings with a large number of clients. Moreover, when the number of clients is exceedingly large, it may be beneficial to cluster clients with similar data distributions and learn a model for each such cluster, ultimately allowing better utilization of the available resources.

\section{Theoretical Analysis}

In this section, we present analytical results that provide insights into Fed-QSSL, with a focus on the impact of low-bitwidth training on the convergence and robustness. For the sake of tractability, we consider the SSL formulation utilizing local objective defined as $\mathcal{L}_{SSL,k}(\w) = -\mathbb{E}[(\w(x_{k, i}) +\xi_{k, i})^T(\w(x_{k, i}) +\xi'_{k, i})] + \frac{1}{2} \|\w^T \w \|^2 $, where $\xi_{k, i}$ and $\xi'_{k, i}$ denote random noise added to the data sample, while the global objective is defined as $\mathcal{L}_{SSL} = \sum_{k=1}^n \frac{|D_k|}{|D|}\mathcal{L}_{SSL, k}(\w)$.
This objective is obtained from the InfoNCE loss $\mathcal{L}_{CL, k}$ by replacing normalization via negative signals by an alternative regularization term \cite{wang2022does}. Optimizing $\mathcal{L}_{SSL}$ is equivalent to minimizing $\mathcal{L}(\w) = \|\bar{X} - \w^T \w \|^2$ where $\bar{X} = \sum_k \frac{|D_k|}{|D|}X_k$ and $X_k = \mathbb{E}_{x \sim D_k}(xx^T) = \frac{1}{|D_k|}\sum_{i=1}^{|D_k|} x_{k, i}x_{k, i}^T$, the empirical covariance matrix of client $k$'s data. Since the aim of this analysis is to assess the impact of low-bitwidth operations at client devices on the convergence and robustness of federated learning, we simplify operations at the server by replacing the sophisticated de-quantizer in Algorithm \ref{alg:algorithm2} by a layer-wise codebook mapping low-bitwidth weights to floating-point values according to $\w'^{(Q)}_{k, t} = \w^{(Q)}_{k, t}$. The subsequent aggregation follows FedAvg and computes $\w_{t, G} = \sum_{k=1}^n \frac{|D_k|}{|D|} \w'^{(Q)}_{k, t}$, while re-quantization uses a $\tanh$ based quantizer.

For simplicity, we assume that clients quantize weights/activations using the same bitwidth, i.e., $s_1 = s_2 = \cdots = s_n = R$; moreover, local gradient quantization is performed using the same number of bits. When bitwidth varies across client devices, the devices with higher bitwidth typically have smaller quantization error. Our forthcoming analysis can be viewed as providing error bounds for a system in which clients deploy bitwidth no less than $R$. Local datasets are generated in a distributed and $2n$-way manner, i.e., local dataset at client $k$ is generated such that the labels of data samples are skewed to classes $2k-1$ and $2k$, with a few samples coming from other classes. This distribution is close to the pathologically non-iid case with two dominant classes. More details about heterogeneous data generation are provided in Appendix.

We consider three types of quantization errors: $\epsilon_{\w}$ denotes the model parameter quantization error, $\epsilon_g$ denotes the gradient quantization error, and $\epsilon_r$ denotes the quantization error induced by the server when re-quantizing the aggregated model. Below we use subscripts $k$ and $t$ to indicate the client and the epoch, respectively. Quantization can be viewed as adding noise to the full precision values, i.e., $\g^{(Q)}_{k, t} = \g_{k, t} + \epsilon_{g_k, t} $. In quantized training at client $k$, the update can be expressed as
\begin{align*}
    \w_{k, t+1}^{(Q)} = \w_{k, t}^{(Q)} - \alpha_t (\g_{k, t} + \epsilon_{g_k, t} )+ \epsilon_{\w_k, t}
\end{align*}
where $\w_{k, t}^{(Q)}$ denotes the low-bitwidth model parameters at time $t$, $\epsilon_{g_k, t}$ is the noise added to the gradient in the quantization step, and $\epsilon_{\w_k, t}$ denotes the noise added to the model parameters after gradient update as the parameters retain low-bitwidth representation. After $E$ local training epochs, the server collects the models and aggregates them according to 
\begin{align*}
    & \w_{t+E, G} = \sum_{k=1}^n \frac{|D_k|}{|D|} \w_{t+E, k}^{(Q)} = \sum_{k=1}^n \frac{|D_k|}{|D|} \w_{k, t}^{{Q}} \\
    & \quad - \sum_{s=0}^{E-1} [\alpha_{t+s}(\g_{k, t+s} + \epsilon_{g_k, t+s} ) - \epsilon_{\w_k, t+s}].
\end{align*}
Since the aggregated model is not necessarily in low bitwidth, an additional re-quantization step is required to form its quantized version $\w_{t+E, G}^{(Q)} = \w_{t+E, G} + \epsilon_{r, t+E}$, which is then sent to the clients.

Quantization centers used for layer-wise scalar quantization of the model parameters and gradients are found via companding quantization. Specifically, full-precision input $\x$ is transformed by a nonlinear function $c$, e.g., $\tanh$ function, and then uniformly quantized. The output $Q(\x)$ is generated by taking the value of the inverse function, $c^{-1}$, of the quantized value \cite{sun2011scalar}. Uniform quantization is  a special case of companding quantization obtained by setting $c(\x) = \x$. In Fed-QSSL, local quantizers can also be viewed as special cases of the companding quantizers.

To proceed with the analysis, we make the following assumption on the quantizers.
\begin{assumption}\label{Assumption-1}
    All quantization operators in the low-bit training are unbiased.
\end{assumption}
This assumption is commonly encountered in prior work (see, e.g., \cite{reisizadeh2020fedpaq}); an example is the stochastic quantizer, i.e., given a sequence of quantization centers $Q_1 \leq \cdots \leq Q_{2^R}$, for $x \in [Q_j, Q_{j+1}]$ we have $Q(x) = Q_j$ with probability $\frac{Q_{j+1} - x}{Q_{j+1} - Q_j}$ and $Q(j) = Q_{j+1}$ with probability $\frac{x - Q_{j}}{Q_{j+1} - Q_j}$. This assumption implies that quantization errors $\epsilon_{\w}$, $\epsilon_{g}$, and $\epsilon_r$ are zero-mean.

We further make the following assumption on gradient estimates, frequently encountered in literature.
\begin{assumption}\label{Assumption-2}
Expected gradient estimate is unbiased and bounded, $\mathbb{E}_t \|\g_{k, t} \|^2 \leq G^2$.
\end{assumption}

The following lemma provides a bound on the quantization errors (for proof please see Appendix).
\begin{lemma}\label{lemma1}
Suppose Assumptions \ref{Assumption-1}-\ref{Assumption-2} hold, and that client $k$ computes update of the quantized model parameters $\w_{k, t+1}^{(Q)}$ at bitwidth
$s_k = R$. Then the gradient quantization error $\epsilon_{g_k, t}$ satisfies $\mathbb{E}_t[\|\epsilon_{g_k, t} \|^2] = \O(G^2/2^{2R})$, and the local re-quantization error $\epsilon_{\w_k, t}$ satisfies $\mathbb{E}_t[ \|\epsilon_{\w_k, t}\|^2] = \O(\alpha_t G^2/2^{2R})$.
\end{lemma}

This lemma indicates that by viewing the gradient update $\alpha_t (\g_{k, t} + \epsilon_{g_k, t} )$ as a perturbation of $\w^{(Q)}_{k, t}$, the variance of the weight quantization error $\epsilon_{\w_k, t}$ in the local update is of the order $\O(\alpha_tG^2)$. It further implies that the variance of the re-quantization error $\epsilon_{r, t}$ in the global update is of the order $\O(\alpha_t G^2 E)$ when $\alpha_{t-E+1} = \cdots = \alpha_t$. 

The next corollary readily follows from Lemma~1.

\begin{corollary}\label{corollary1}
Instate the assumptions of Lemma \ref{lemma1}, and let all clients use the same learning rate $\alpha_{t-E+1} = \cdots = \alpha_t$. Then there exists $G_q = \O(G^2/2^{2R})$ such that $\mathbb{E}_t[\|\epsilon_{\w_k, t} \|^2 ]\leq \alpha_t G_q^2 $ and $\mathbb{E}_t[\|\epsilon_{r, t} \|^2 ]\leq \alpha_t G_q^2 $.
\end{corollary}

Next, we consider the convergence of the quantized SSL training. Note that objective $\mathcal{L} = \|\bar{X} - \w^T \w \|^2$ is generally a non-convex and non-smooth function. In our analysis, we consider a class of functions that satisfies the $\rho$-weak convexity. 

\begin{definition}
A function $f: \R^d \to \R $ is $\rho$-weakly convex if $f(x) + \frac{\rho}{2}\|x \|^2$ is convex.
\end{definition}
Generally, any function of the form $f(x) = g(h(x))$, where $g(\cdot)$ is convex and Lipschitz and $h(\cdot)$ is a smooth map with Lipschitz Jacobian, is weakly-convex.
Clearly, the federated SSL objective satisfies $\rho$-weak convexity with $\rho \geq 4\|\bar{X} \|$. To analyze weakly-convex objective function $\mathcal{L}$, we introduce the Moreau envelope
\begin{align*}
    \phi_{\lambda}(x) := \min_y \{\mathcal{L}(y) + \frac{1}{2\lambda}\|y-x \|^2 \}
\end{align*}
and define the corresponding proximal map as
\begin{align*}
    \mathrm{prox}_{\phi_{\lambda}(x) } : = \mathrm{argmin}_y  \{\mathcal{L}(y) + \frac{1}{2\lambda}\|y-x \|^2 \}.
\end{align*}

We use $\|\nabla \phi_{\lambda}(x) \|$ as the convergence indicator. Intuitively, small norm of the gradient of $\phi_{\lambda}(x) $ implies that $x$ is close to a point $\hat{x}$ that is stationary for $\phi$. For Fed-QSSL, the following result holds.

\begin{theorem}\label{thm1}
Suppose all assumptions of Lemma \ref{lemma1} and Corollary \ref{corollary1} hold.
For all $\bar{\rho} > \rho$, after $T$ communication rounds of Fed-QSSL
\begin{align*}
    & \frac{1}{\sum_{t=0}^T \alpha_{tE}} \sum_{t=0}^T \alpha_{tE} \mathbb{E} [ \|\nabla \phi_{\frac{1}{\bar{\rho}}} (\w^{(Q)}_{tE, G}) \|^2 ] \leq  \frac{E \bar{\rho}}{\bar{\rho} - \rho} \times \\& \frac{ \phi_{\frac{1}{\bar{\rho}}}(\w^{(Q)}_{0, G}) - \min \phi +  \bar{\rho}(G^2\sum_{t=0}^T \alpha^2_{tE}+3G_q^2\sum_{t=0}^T \alpha_{tE}) }{\bar{\rho}\sum_{t=0}^T \alpha_{tE} },
\end{align*}
where $E$ denotes the number of local training epochs per communication round (for a total of $tE$ training epochs after $t$ communication rounds) and $\w^{(Q)}_{0, G}$ is the quantized parameter initialization. 
\end{theorem}
Theorem~1 implies convergence of the algorithm to a nearly stationary state of the objective function. When $\alpha_t$ decreases as $\O(1/\sqrt{t})$, the upper bound vanishes as $t \to \infty$. The optimal solution $\w^*$ of the global objective satisfies $\| \nabla \phi_{\frac{1}{\bar{\rho}}} (\w^*) \|^2  = 0$; the vanishing upper bound implies proximity to a neighborhood of the optimal solution.{\footnote{While using gradient descent estimates with bounded variance typically leads to converge to local minima \cite{mertikopoulos2020almost}, a property of the SSL objective is that all local minima are global minima \cite{jin2017escape}.}} As the quantization rate $R$ approaches full precision rate, $G_q$ vanishes and the upper bound collapses to the value obtained by vanilla stochastic gradient descent analysis.

It is further of interest to analyze robustness of the learned representations. To this end, we first formally define the representation vector.
\begin{definition}\cite{wang2022does}
    Let $\mathcal{S} \subset \R^d$ be the subspace spanned by the rows of the learned feature matrix $\w \in \R^{m \times d}$ for the embedding function $f_{\w}(x) = \w x$. The representability of $\mathcal{S}$ is defined as the vector $r = [r_1, \cdots, r_d]^T$ such that $r_i = \|\Pi_{\mathcal{S}}(e_i) \|^2$ for $i \in [d]$, where $\Pi_{\mathcal{S}}(e_i) $ denotes the projection of standard basis $e_i$ onto $\mathcal{S}$ and thus $r_i = \sum_{j=1}^s\langle e_i, v_j \rangle^2$, $s = \dim(\mathcal{S})$ and $\{v_j \}$ is the set of orthonormal bases of $\mathcal{S}$.
\end{definition}

Vector $r$ introduced in this definition quantifies representability by comparing unit bases of different feature spaces. A good feature space should be such that the entries of $r$ are large, indicating its standard unit bases can well represent $e_1, \dots, e_d$, the unit bases in $\R^d$. 

\begin{theorem}\label{thm2}
Suppose assumptions of Theorem \ref{thm1} hold and $n = \Theta(d^{1/20})$. In a $2n$-way classification task, when in local training the update is $\epsilon$ away from the optimal solution $\w_k^*$, the representation vector learned by client $k$ with high probability satisfies $ \frac{d^{2/5} - \O(d^{-2/5}) + 2e_j^T (\w_k^*)^T \epsilon e_j + (e_j^T \epsilon)^2}{d^{2/5} + \O(d^{-2/5}) + \|2(\w_k^*)^T \epsilon + \epsilon^T \epsilon  \|} \leq r_{i}^k \leq 1$ for $i \in [n]\backslash \{k \}$. As for the global objective, when the update is $\epsilon_1$ away from the optimal solution $\w^*$, the learned representation vector $\bar{r}$ with high probability satisfies $\frac{d^{2/5} - \Theta(d^{7/20}) + \O(d^{-1/20}) - \O(d^{2/5}) + 2e_j^T (\w^*)^T \epsilon_1 e_j + (e_j^T \epsilon_1)^2 }{d^{2/5} - \Theta(d^{7/20}) + \O(d^{-1/20}) + \|2(\w^*)^T \epsilon_1 + \epsilon_1^T \epsilon_1  \|} \leq \bar{r}_i \leq 1$ for all $i \in [n]$. 
\end{theorem}

Theorem \ref{thm2} implies that as $\epsilon_1$ vanishes, the entries in representation vectors have strictly positive lower bound and thus do not vanish. This theorem states that the representation space learned from the SSL objectives is such that for the $2n$ basis directions that generate the data, any two clients have similar representability.
The theorem further implies that the learned representation vectors are not biased towards local data distributions and are capable of performing well on uniformly distributed data, indicating robustness of the scheme. 
For the proof of Theorem~2, please see Appendix.


\section{Experimental Results}

We simulate an FL system with $10$ clients, where the participating clients have different bitwidth configuration and local data distributions. We deploy two bitwidth configurations: (1) the configuration with gradually increasing bitwidth: $20\% 4$-bit, $30\% 6$--bit, $30\% 8$-bit and $20\% 12$-bit models; and (2) the configuration with a skewed bitwidth: $50\% 6$-bit and $50\% 12$-bit models. The experiments involve CIFAR-10 and CIFAR-100 datasets, both for image classification tasks \cite{krizhevsky2009learning}. The datasets are distributed to clients in a non-iid fashion according to Dirichlet$(0.1)$ distribution \cite{bibikar2022federated}.
Simulations were executed on AMD Vega $20$ GPUs. The learning rates were set as in the following prior work: for the supervised learning schemes using low-bit operations we deploy learning rates as in \cite{zhou2016dorefa}, while for self-supervised learning schemes we follow \cite{wang2022does}.
In all simulations, the deployed neural networks were based on ResNet-18 architecture \cite{he2016deep}.
The results are reported after $100$ communication rounds, with clients running $E = 20$ local epochs between any two rounds in CIFAR-10 simulations and $E = 5$ in CIFAR-100 simulations. In the implementation of Fed-QSSL, we rely on the SimCLR method \cite{chen2020simple} for the self-supervision part of the algorithm.\footnote{The codes are available at \url{https://github.com/YiyueC/Fed-QSSL}.}

\begin{table}[t]
\centering
\begin{tabular}{c|c|c}
    \hline
    Algorithm & Global Acc & Local Acc \\
    \hline
    FedAvg & 17.70 & 64.98  \\
    \hline
    FedProx & 14.51 & 68.46  \\
    \hline
    FedPAQ & 16.68 & 62.96  \\
    \hline
    Fed-SimCLR & 39.09 & 77.30  \\
    \hline
    Fed-SimSiam & 36.73 & 76.87 \\
    \hline
    \textbf{Fed-QSSL} & \textbf{59.31} & \textbf{82.50} \\
    \hline 
    \hline
    Fed-SimCLR (Full) & 40.76 & 80.56 \\
    \hline
    \textbf{Fed-QSSL} (Full) & \textbf{72.26} & \textbf{90.01} \\
    \hline
\end{tabular}
\caption{Experiments on CIFAR-10 with model bitwidth configuration $20\% 4$-bit, $30\% 6$--bit, $30\% 8$-bit and $20\% 12$-bit.}
\label{table-1}
\end{table}

\begin{table}[t]
\centering
\begin{tabular}{c|c|c}
    \hline
    Algorithm & Global Acc & Local Acc \\
    \hline
    FedAvg & 20.77 & 66.64  \\
    \hline
    FedProx & 18.70 & 75.60  \\
    \hline
    FedPAQ & 18.56 & 76.57 \\
    \hline
    Fed-SimCLR & 39.12 & 77.43  \\
    \hline
    Fed-SimSiam & 36.44 & 75.21 \\
    \hline
    \textbf{Fed-QSSL} & \textbf{62.26} & \textbf{83.03} \\
    \hline \hline
    Fed-SimCLR (Full) & 40.76 & 80.56 \\
    \hline
    \textbf{Fed-QSSL} (Full) & \textbf{72.26} & \textbf{90.01} \\
    \hline
\end{tabular}
\caption{Experiments on CIFAR-10 with model bitwidth configuration $50\% 6$-bit and $50\% 12$-bit.}
\label{table-2}
\end{table}

\begin{table}[t]
\centering
\begin{tabular}{c|c|c}
    \hline
    Algorithm & Global Acc & Local Acc \\
    \hline
    Fed-SimCLR & 14.79 & 23.46  \\
    \hline
    Fed-SimSiam & 12.24 & 13.59 \\
    \hline
    \textbf{Fed-QSSL} & \textbf{21.44} & \textbf{30.91} \\
    \hline \hline
    Fed-SimCLR (Full) & 15.18 & 30.26 \\
    \hline
    \textbf{Fed-QSSL} (Full) & \textbf{28.75} & \textbf{43.56} \\
    \hline
\end{tabular}
\caption{Experiments on CIFAR-100 with bitwidth configuration $50\% 6$-bit and $50\% 12$-bit.}
\label{table-3}
\end{table}

\begin{figure}
\centering
    \begin{subfigure}
  \centering
  \includegraphics[width=.8\linewidth]{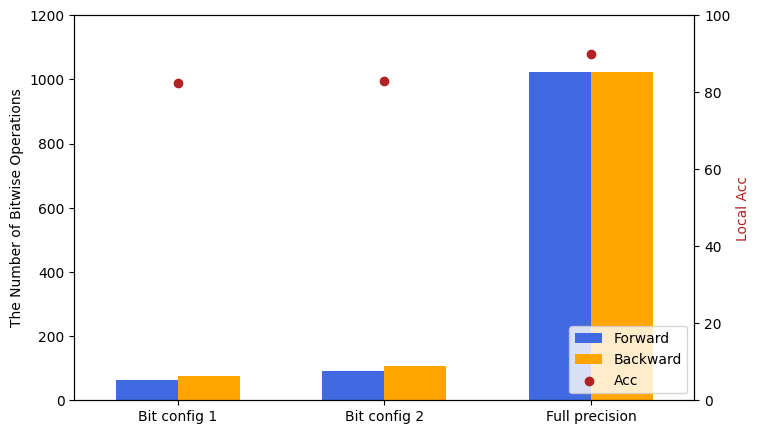}
  \caption{The left y-axis indicates the number of bitwise operations while the right y-axis indicates the local accuracy achieved by Fed-QSSL.}
  \label{fig:sfig1}
\end{subfigure}
\vspace{2mm}
\begin{subfigure}
  \centering
  \includegraphics[width=.8\linewidth]{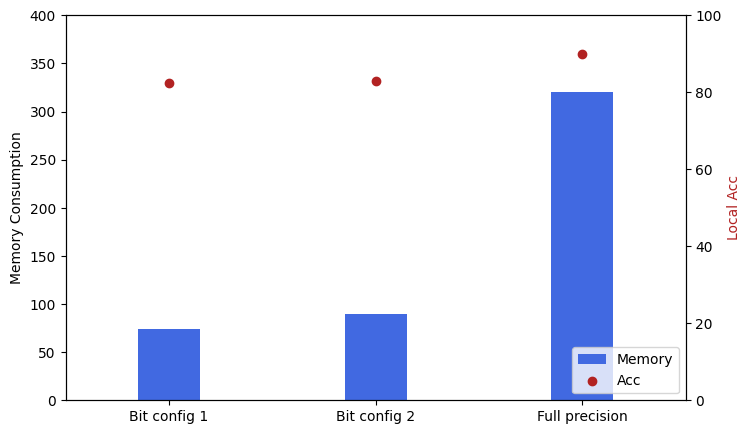}
  \caption{The left y-axis indicates the memory consumption while the right y-axis indicates the local accuracy achieved by Fed-QSSL.}
  \label{fig:sfig2}
\end{subfigure}
\end{figure}

\begin{figure*}
\begin{minipage}[t]{0.3\textwidth}
  \includegraphics[width=\linewidth]{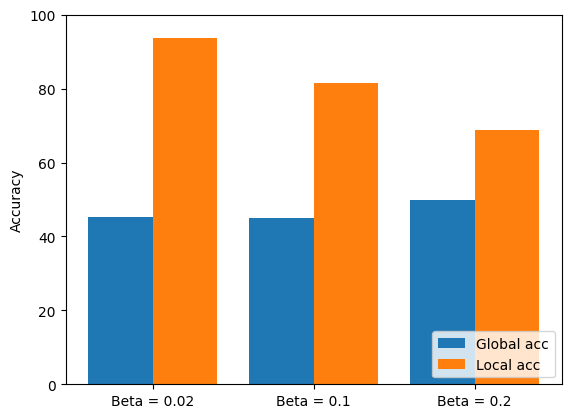}
  \caption{Ablation study on Dirichlet distribution parameter $\beta$.}
  \label{fig:first}
\end{minipage}%
\hfill 
\begin{minipage}[t]{0.3\textwidth}
  \includegraphics[width=\linewidth]{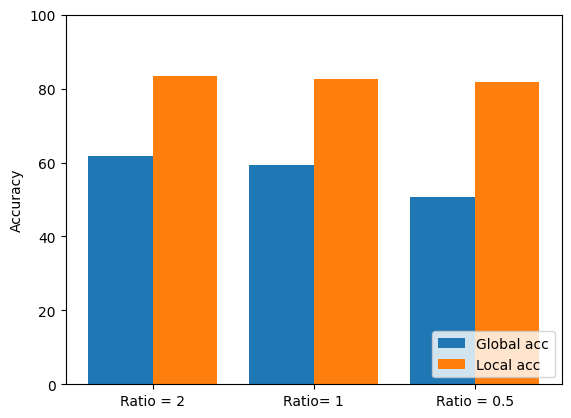}
  \caption{Ablation study on the ratio of the global buffer size and the local dataset size.}
  \label{fig:second}
\end{minipage}%
\hfill
\begin{minipage}[t]{0.3\textwidth}
  \includegraphics[width=\linewidth]{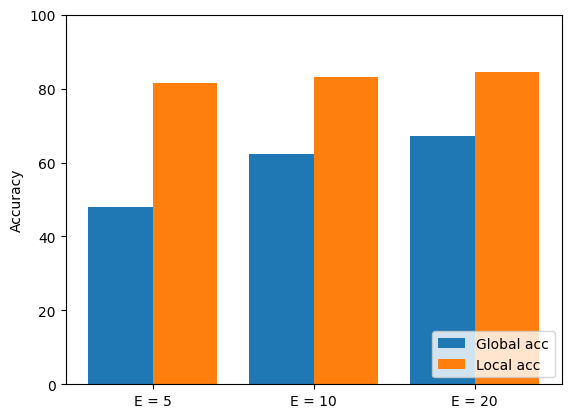}
  \caption{Ablation study on the number of local training epochs $E$.}
  \label{fig:third}
\end{minipage}%
\end{figure*}

{\bf Baselines.} We compare Fed-QSSL with two groups of algorithms: federated supervised learning (SL) and self-supervised algorithms. The SL algorithms include FedAvg \cite{mcmahan2017communication}, classic FL technique performing simple averaging of local models at each communication round; FedProx \cite{li2020federated}, an FL scheme addressing client data heterogeneity by adding an $\ell_2$-norm regularizer to local objectives to prevent divergence of local updates from the global model; and FedPAQ \cite{reisizadeh2020fedpaq}, a communication-efficient FL algorithm where clients transmit quantized updates to reduce uplink communication cost. As for the SSL algorithms, we consider Fed-SimCLR \cite{chen2020simple,wang2022does} which uses contrastive learning objective to learn a global feature extraction model; and Fed-SimSiam \cite{chen2021exploring} which considers only positive pairs of data points and learn meaningful features by leveraging a feature predictor function and stop-gradient operation. While the baseline algorithms are not originally meant to support local low-bit training, we apply low-bitwidth operations to satisfy resource constraints. In each table of results, the last two rows (labeled as ``Full") correspond to full precision ($32$-bit) models, i.e., models with no bitwidth constraints. 

{\bf Metrics.} Fed-QSSL is compared to the baselines in terms of global accuracy (Global Acc), reflective of robustness, and local accuracy (Local Acc), reflecting personalization. 
SSL methods train a universal linear classifier on top of the frozen global feature extraction model and evaluate the classification accuracy on a uniformly distributed test dataset. Training of a linear classifier to be used for testing global model accuracy is conducted at full precision on a centralized training dataset as in \cite{wang2022does}; this is done only to validate robustness of SSL methods, in real applications no such universal linear classifier is required. SSL methods use the aggregated model and test on uniform testing dataset.
To evaluate local accuracy (i.e., the accuracy on local, generally non-IID datasets), SSL methods train a local linear classifier on top of the frozen low-bitwidth feature extractor while meeting client's bitwidth constraints and run it on a local non-iid testing dataset, while SL methods use a local low-bitwidth classification model and compute the accuracy on local testing data.

{\bf Results.} The results on CIFAR-10 under the first bitwidth configuration are reported in Table \ref{table-1}. Overall, the global accuracy of SSL schemes is higher than SL methods, implying that SSL algorithms tend to learn more accurate representations in data heterogeneous FL scenario. As for the personalization, SSL algorithms achieve higher accuracy on local data, suggesting they learn meaningful representations that are then used in the downstream classification task. It is worth noting that SSL algorithms with higher global accuracy also achieve higher local accuracy; this is because the learned robust feature extractor extracts expressive representations from heterogeneous data which help in downstream classification tasks. Among SL algorithms, however, there exists a trade-off between robustness and personalization, where the performance of models with high global accuracy may suffer locally. Specifically, Fed-QSSL performs better in both bitwidth-heterogeneous and no-constrained (full precision) scenarios due to the use of server-side operations on the global buffer; operations on server facilitate learning robust representations that further improve local performance. Table \ref{table-2} reports results on CIFAR-10 under the second bitwidth configuration. There, SSL algorithms again demonstrate more robust and personalized performance in face of bitwidth and data heterogeneity. As the permitted bitwidth grows, the algorithms reach higher accuracy. Nevertheless, under the bitwidth constraints Fed-QSSL still achieves the highest accuracy. Finally, Table \ref{table-3} reports results on the more challenging CIFAR-100 dataset. As can be seen there, Fed-QSSL perform better than other SSL benchmarks under both low-bitwidth as well as full precision scenarios.


{\bf Computation and Memory Saving.}
We next compare the computational cost and memory requirements of Fed-QSSL to those of the full precision scheme. In particular,
for the two bitwidth configurations and the full precision scenario considered above, we report the number of bitwise operations and memory footprint in the local training (see Figure \ref{fig:sfig1} and Figure \ref{fig:sfig2}). In Figure \ref{fig:sfig1}, we show the number of bitwise operations for the considered bitwidth configurations; as can be seen there, in both forward and backward pass the computational cost of Fed-QSSL is much smaller than that of the full precision scheme. In low-bit configurations, the backward propagation is consuming more computation because the gradient are represented using 2 bits more than the weights. Figure \ref{fig:sfig2} shows that Fed-QSSL achieves significant memory savings while still providing competitive accuracy.

{\bf Ablation Study.}
Lastly, Figures \ref{fig:first}-\ref{fig:third} present results of ablation studies. When increasing Dirichlet distribution parameter $\beta$, local data distributions become more uniform and the global accuracy increases. The local accuracy, however, decreases since the clients start facing more classes and thus need to execute more challenging classification tasks (Fig.~3). When smaller size global buffer is used and the ratio of the size of $D_g$ and the size of local data decreases, the global accuracy deteriorates while the local accuracy remains mostly unchanged (Fig.~4). Finally, as more training epochs are used in local training, Fed-QSSL achieves higher accuracy (Fig.~5).

\section{Conclusion and Future Work}
We introduced the federated quantized self-supervised learning (Fed-QSSL) algorithm, an effective framework for FL in bitwidth and data heterogeneity settings. We analytically studied the impact of low-bit training on the convergence and robustness of FL, and experimentally demonstrated that Fed-QSSL achieves more robust and personalized performance than benchmarking algorithms. Future work may include an extension to large-scale settings where it is meaningful to cluster clients with similar data distributions, and train per-cluster models. 
For such systems, it is of interest to develop schemes that aim to optimally manage utilization of the available resources.

	\section*{A. Appendix}
	This section is organized as follows. Section~A.1 provides notation and sets up the stage for theoretical analysis. Section A.2-4 present proofs of Lemma~1, Theorem~1 and Theorem~2, respectively. Section~A.5 contains details of the local training under low bitwidth constraints.
	
	\subsection*{A.1. Preliminaries}
	\subsubsection*{Notation}
Let $n$ denote the number of clients in a FL system. Client $k$ owns dataset $D_k$ containing $|D_k|$ samples drawn independently from distribution $\mathcal{D}_k$. Data sample $x \in \R^d$ is a $d$-dimensional vector. Parameters of the learned SSL model are denoted by $\w \in \R^{m \times d}$. We use $[n]$ to denote the set $\{1, 2, \cdots, n \}$. Let $\mathcal{L}$ denote the global objective function, while $\mathcal{L}_k$ denotes local objective at client $k$. The inner product of two vectors, $\x$ and $\z$, is denoted by $\langle \x, \z \rangle$. Finally, let $\|\cdot \|$ be the $\ell_2$-norm of a vector or Frobenius norm of a matrix, and let $\| \cdot \|_2$ denote the $\ell_2$-norm of a matrix.

\subsubsection*{Background and settings}


We consider a federated learning system with $n$ clients participating in model learning via self-supervised learning (SSL). The local objective function used by client $k$ is
\begin{align*}
    \mathcal{L}_{SSL, k}(\w) = -\mathbb{E}[(\w(x_{k, i}) +\xi_{k, i})^T(\w(x_{k, i}) +\xi'_{k, i})] + \frac{1}{2} \|\w^T \w \|^2,
\end{align*}
where $\xi_{k, i}$ and $\xi'_{k, i} $ denote random noise added to the samples. Then the global objective function is defined as $\mathcal{L}_{SSL} = \sum_{k=1}^n \frac{|D_k|}{|D|}\mathcal{L}_{SSL, k}(\w)$. The goal of the FL system is to find $\w$ that minimizes the global objective. The optimization problem can be written equivalently as $\min \mathcal{L}(\w) = \|\bar{X} - \w^T \w \|^2$, where $\bar{X} = \sum_k \frac{|D_k|}{|D|}X_k$ and $X_k = \mathbb{E}_{x \sim D_k}(xx^T) = \frac{1}{|D_k|}\sum_{i=1}^{|D_k|} x_{k, i}x_{k, i}^T$ denotes the empirical data covariance matrix at client $k$. Minimization of the local objective can equivalently be stated as $\min \mathcal{L}_{k}(\w) = \|X_k - \w^T \w\|^2$. 

Recall that the focus of our analysis is on the impact of low-bit operations deployed in local training on the convergence and robustness of Fed-QSSL. For analytical tractability, we simplify operations at the server by replacing sophisticated de-quantizer by a layer-wise codebook mapping low-bitwidth weights to floating-point values according to $\w'^{(Q)}_{k, t} = \w^{(Q)}_{k, t}$. The subsequent aggregation follows FedAvg and computes $\w_{t, G} = \sum_{k=1}^n \frac{|D_k|}{|D|} \w'^{(Q)}_{k, t}$, while re-quantization uses the $\tanh$-based quantizer. For simplicity, we assume that the clients quantize weights/activations using the same bitwidth, i.e., $s_1 = s_2 = \cdots = s_n = R$; moreover, local gradient quantization is performed using the same number of bits (i.e., $s_k$).

Quantization centers used for layer-wise scalar quantization of the model parameters and gradients are found via companding quantization. Specifically, full-precision input $\x$ is transformed by a nonlinear function $c$, e.g., $\tanh$ function, and then uniformly quantized. The output $Q(\x)$ is generated by taking the value of the inverse function, $c^{-1}$, of the quantized value \cite{sun2011scalar}. Uniform quantization is  a special case of companding quantization, obtained by setting $c(\x) = \x$. In Fed-QSSL, local quantizers can also be viewed as special cases of the companding quantizers. We consider fixed rate quantization, where each codeword (quantization center) used for quantization has the same length. When the quantization rate is $R$, the codebook has $K = 2^R$ codewords \cite{sun2011scalar}. 

We further assume that the quantization noise has zero mean, which can be achieved by adopting stochastic quantization: Given a sequence of quantization centers $Q_1 \leq \cdots \leq Q_{2^R}$, for $x \in [Q_j, Q_{j+1}]$ we have $Q(x) = Q_j$ with probability $\frac{Q_{j+1} - x}{Q_{j+1} - Q_j}$ and $Q(j) = Q_{j+1}$ with probability $\frac{x - Q_{j}}{Q_{j+1} - Q_j}$.

    \subsection*{A.2. Proof of Lemma~1}
    We first review notation featured in the lemma. 
Let $\epsilon_{\w}$ and $\epsilon_g$ denote the error induced by quantizing parameters and gradients, respectively. Let $\epsilon_r$ be the quantization error induced by the server when re-quantizing an aggregated model. 

When training at full precision using gradient descent, assuming $E$ epochs of local training per communication round, the global model update is formed as
\begin{align*}
    \w_{t+E, G} = \sum_{k=1}^n \frac{|D_k|}{|D|}(\w_{t, G} - \sum_{s = 0}^{E-1} \alpha_{t+s} \frac{\partial \mathcal{L}_{SSL, k}}{\partial \w_{t+s, k}}).
\end{align*}

When performing low-bit quantized training, quantized weights of the model trained by client $k$ are found as 
\begin{align*}
    \w_{k, t+1}^{(Q)} = \w_{k, t}^{(Q)} - \alpha_t (\g_{k, t} + \epsilon_{g_k, t} )+ \epsilon_{\w_k, t},
\end{align*}
where $\w_{k, t}^{(Q)}$ denotes the low-bitwidth model parameters at time $t$, $\epsilon_{g_k, t} $ is the noise added to the gradient in the quantization step, and $\epsilon_{\w_k, t}$ denotes the noise added to the model parameters after gradient update as the parameters retain low bitwidth representation. After $E$ local training epochs, the server collects the models and aggregates them according to 
\begin{eqnarray*}
    \w_{t+E, G} & = &\sum_{k=1}^n \frac{|D_k|}{|D|} \w_{t+E, k}^{(Q)} \\ & = & \sum_{k=1}^n \frac{|D_k|}{|D|} \w_{k, t}^{{Q}}
    - \sum_{s=0}^{E-1} [\alpha_{t+s}(\g_{k, t+s} + \epsilon_{g_k, t+s} ) - \epsilon_{\w_k, t+s}].
\end{eqnarray*}
Since the aggregated model is not necessarily in low bit, an additional re-quantization step is required to form its quantized version $\w_{t+E, G}^{(Q)} = \w_{t+E, G} + \epsilon_{r, t+E}$, which is then sent to clients.

We proceed to Lemma~1, an intermediate result characterizing the variance of quantization errors, $\mathbb{E}_t[\|\epsilon_{g_k, t} \|^2]$ and $\mathbb{E}_t[ \|\epsilon_{\w_k, t}\|^2] $. The lemma shows that given low-bitwidth model parameters and small gradient perturbation, the variance of $\epsilon_w$ is proportional to the norm of the gradient perturbation. In other words, when a small perturbation is added to low-bitwidth model parameters, the noise induced by quantizing the perturbed parameters has small variance.

For the result in Lemma~1, we need to make the following assumptions.

\newtheorem*{assumption1}{Assumption 1}
\begin{assumption1}
    All quantization operators in the low-bit training are unbiased.
\end{assumption1}
\newtheorem*{assumption2}{Assumption 2}
\begin{assumption2}
    Expected gradient estimate is unbiased and bounded, 
    
    $\mathbb{E}_t \|\g_{k, t} \|^2 \leq G^2$.
\end{assumption2}

\newtheorem*{lemma1}{Lemma 1}
\begin{lemma1}
Suppose Assumptions \ref{Assumption-1}-\ref{Assumption-2} hold, and that client $k$ computes update of the quantized model parameters $\w_{k, t+1}^{(Q)}$ at bitwidth
$s_k = R$. Then the gradient quantization error $\epsilon_{g_k, t}$ satisfies $\mathbb{E}_t[\|\epsilon_{g_k, t} \|^2] = \O(G^2/2^{2R})$, and the local re-quantization error $\epsilon_{\w_k, t}$ satisfies $\mathbb{E}_t[ \|\epsilon_{\w_k, t}\|^2] = \O(\alpha_t G^2/2^{2R})$.
\end{lemma1}

\begin{proof}
Recall that the quantization in Algorithm~1 is conducted layer-wise. Thus, it suffices to analyze quantization errors for the parameters in the first layer -- the results readily extend to the parameters in the remaining layers. Given a local update
\begin{align*}
    \w_{k, t+1}^{(Q)} = \w_{k, t}^{(Q)} - \alpha_t (\g_{k, t} + \epsilon_{g_k, t} )+ \epsilon_{\w_k, t},
\end{align*}
the gradient update can be viewed as an $\O(\alpha_t \|\g_{k, t}\|)$ perturbation of the corresponding original codebook value. Since $\mathbb{E}_t\|\g_{k, t} \|^2 \leq G^2$ and we use quantile quantization for gradient compression, the gradient quantization error $\epsilon_{g_k, t}$ has range within $\O(\|\g_{k, t}\|)$. 
Therefore, variance of the gradient quantization error $\mathbb{E}_t[\|\epsilon_{g_k, t}\|^2]$ is $\O(G^2)$. 

To analyze weight quantization, let $\w_{\min}$ and $\w_{\max}$ be the minimal and maximal value in $\w_{k, t}^{(Q)}$, respectively, and let $\w'_{\min}$ and $\w'_{\max}$ denote the minimal and maximal value in $\w_{k, t}^{(Q)} - \alpha_t (\g_{k, t} + \epsilon_{g_k, t} )$, respectively. It then holds that $| \w_{\min} - \w'_{\min} | $ is $\O(\alpha_t \|\g_{k, t}\|)$ and $| \w_{\max} - \w'_{\max} | $ is $\O(\alpha_t \|\g_{k, t}\|)$. By leveraging fixed rate companding quantization with $\tanh$ compander and the fact that the derivative of $\tanh$ decreases from $1$ to $0$ as the absolute value of the variable increases, it can be shown that the $k$-th codeword in the codebook for $\w^{(Q)}_{k, t+1}$ is at most $\O(\alpha_t \|\g_{k, t}\|)$ away from the $k$-th codeword in the codebook for $\w^{(Q)}_{k, t}$, which implies that by using stochastic quantization operator, the variance of $\epsilon_{\w_k, t}$, $\mathbb{E}_t[\|\epsilon_{\w_k, t}\|^2 ]$, is $\O(\alpha_t G^2)$.


In order to derive the relationship between the variance of the quantization error and the quantization rate in Fed-QSSL, we consider two types of quantizations: uniform quantization and companding quantization with optimal compander. Since neural network weights are typically not uniformly distributed but rather follow centered distributions such as Gaussian or Laplacian \cite{isik2022information}, the uniform quantization MSE, which is $\O(1/2^{2R})$ with respect to $R$, can be used as an upper bound on the quantization error. Regarding the lower bound, \cite{sun2011scalar} has shown that the lowest MSE for companding quantization is achieved by the optimal compander with the corresponding MSE $\O(1/2^{2R})$ with respect to $R$. Therefore, given the quantization rate $R$, the error of the weight quantizer in Fed-QSSL is $\O(1/2^{2R})$. As for the gradient quantizer which relies on the quantile quantization method, one can invoke the quantization MSE of uniform quantization as an upper bound and the MSE of the optimal companding quantization as a lower bound. By combining the two, it follows that the gradient quantization error is $\O(1/2^{2R})$.
Combining the orders with respect to $G$ and $R$, we can complete the proof.
\end{proof}

The corollary below follows directly from the lemma.

\newtheorem*{corollary1}{Corollary 1}
\begin{corollary1}
Instate the assumptions of Lemma~1, and let all clients use the same learning rate $\alpha_{t-E+1} = \cdots = \alpha_t$ within the same communication round. Then there exists $G_q = \O(G^2/2^{2R})$ such that $\mathbb{E}_t[\|\epsilon_{\w_k, t} \|^2 ]\leq \alpha_t G_q^2 $ and $\mathbb{E}_t[\|\epsilon_{r, t} \|^2 ]\leq \alpha_t G_q^2 $.
\end{corollary1}
\begin{proof}
From Lemma~1, when locally performing single-step gradient update, the variance of $\epsilon_{\w_k, t}$ is $\O(\alpha_t G^2/2^{2R})$. As for the global update, since all clients run $E$ local training epochs, each with constant learning rate, and the update is formed by summing $E$ gradient update, the variance of the re-quantization error $\epsilon_{r, t}$ is $\O(\alpha_t G^2 E) $. Therefore, there exists $G_q =  \O(G^2/2^{2R})$ such that $\mathbb{E}_t[\|\epsilon_{\w_k, t} \|^2 ]\leq \alpha_t G_q^2 $ and $\mathbb{E}_t[\|\epsilon_{r, t} \|^2 ]\leq \alpha_t G_q^2 $.
\end{proof}

    \subsection*{A.3. Proof of Theorem~1}
    We first review several key definitions and then re-state the theorem.
Recall that the global objective $\mathcal{L}$ is non-convex and non-smooth but satisfies weakly-convex assumption. We thus proceed with convergence analysis by relying on techniques encountered in dealing with weakly convex objectives \cite{davis2019stochastic}.

Recall that a function $f: \R^d \to \R $ is $\rho$-weakly convex if $f(x) + \frac{\rho}{2}\|x \|^2$ is convex; this definition is readily extended to matrix variable $\w \in \R^{m \times d}$. In general, any function of the form $f(x) = g(h(x))$, where $g(\cdot)$ is convex and Lipschitz and $h(\cdot)$ is a smooth map with Lipschitz Jacobian, is weakly-convex \cite{davis2019stochastic}.
Since the global objective function is $ \mathcal{L} = \|\bar{X} - \w^T \w \|^2$,
it is straightforward to see that the global SSL objective satisfies weak convexity with $\rho \geq 4\|\bar{X} \|$. For a weakly-convex function $\mathcal{L}$, the key construction is the Moreau envelope,
\begin{align*}
    \phi_{\lambda}(x) := \min_y \{\mathcal{L}(y) + \frac{1}{2\lambda}\|y-x \|^2 \},
\end{align*}
with the corresponding proximal map defined as
\begin{align*}    \mathrm{prox}_{\phi_{\lambda}(x) } : = \mathrm{argmin}_y  \{\mathcal{L}(y) + \frac{1}{2\lambda}\|y-x \|^2 \}.
\end{align*}

The convergence can be quantified and tracked via $\|\nabla \phi_{\lambda}(x) \|$; in particular, small norm of the gradient of $\phi_{\lambda}(x) $ implies that $x$ is close to a point $\hat{x} := \mathrm{prox}_{\phi_{\lambda}}(x)$ which is stationary for $\phi$. 

\newtheorem*{theorem1}{Theorem 1}
\begin{theorem1}
Suppose all assumptions in Lemma~1 and Corollary~1 hold.
For all $\bar{\rho} > \rho$, after $T$ communication rounds of Fed-QSSL
\begin{align*}
    & \frac{1}{\sum_{t=0}^T \alpha_{tE}} \sum_{t=0}^T \alpha_{tE} \mathbb{E} [ \|\nabla \phi_{\frac{1}{\bar{\rho}}} (\w^{(Q)}_{tE, G}) \|^2 ] \\ & \leq \frac{E \bar{\rho}}{\bar{\rho} - \rho} \frac{ \phi_{\frac{1}{\bar{\rho}}}(\w^{(Q)}_{0, G}) - \min \phi +  \bar{\rho}(G^2\sum_{t=0}^T \alpha^2_{tE}+3G_q^2\sum_{t=0}^T \alpha_{tE})}{\bar{\rho}\sum_{t=0}^T \alpha_{tE} },
\end{align*}
where $E$ denotes the number of local training epochs per communication round (for a total of $tE$ training epochs after $t$ communication rounds) and $\w^{(Q)}_{0, G}$ is the quantized parameter initialization. 
\end{theorem1}

\begin{proof}
To analyze the FedAvg-based aggregation with full device participation and $E$ local epochs per communication round, we introduce the following notations. Let $\w^{(Q)}_{k, t}$ denote the parameters of model deployed by client $k$ at time $t$. When $t+ 1 \in \mathcal{I}_E = \{E, 2E, \cdots \}$, the server collects local models and performs aggregation. Let us consider two sequences,
\begin{align*}
    \v_{k, t+1} & = \w^{(Q)}_{k, t} - \alpha_t \g^{(Q)}_k(\w_{k, t}) + \epsilon_{\w_k, t} \\
    \w^{(Q)}_{k, t+1} & = \begin{cases}
    \v_{k, t+1} & t+1 \notin \mathcal{I}_E \\
    \sum_{k=1}^n p_k \v_{k, t+1} + \epsilon_{r, t+1} & t+1 \in \mathcal{I}_E
    \end{cases}
\end{align*}
where the auxiliary variable $\v_{t+1, k}$ is an intermediate result of one-step quantized (stochastic) gradient descent update for $\w^{(Q)}_{k, t}$ \footnote{Please note that this notation is consistent with the global update $\w^{(Q)}_{k, t} $ in Algorithm~1 when $t \in \mathcal{I}_E$. For all other time steps, global aggregation is not accessible and has no bearing to real applications.} and $p_k = \frac{|D_k|}{|D|}$. As stated before, local quantized gradient estimate $\g^{(Q)}_k(\w_{k, t})$ is assumed to be an unbiased estimate of the true gradient. For the global model analysis we examine two virtual sequences, $\{\v_{t, G} \}$ and $ \{ \w^{(Q)}_{t, G} \}$, with $\v_{t, G} = \sum_{k=1}^n p_k \v_{k, t}$ and $\w^{(Q)}_{t, G} = \sum_{k=1}^n p_k\w^{(Q)}_{k, t}$. In the full participation scenario, it always holds that $\v_{t, G} = \w^{(Q)}_{t, G} $ when $t \notin \mathcal{I}_E$ and 
$\v_{t, G} + \epsilon_{r, t} = \w^{(Q)}_{t, G}$ otherwise. 

Define $v_t = \mathbb{E}_t[\g(\w^{(Q)}_{t, G})]$ where $\g(\w^{(Q)}_{t, G}) = \sum_{k=1}^n p_k (\g^{(Q)}_{k, t} - \epsilon_{\w_k, t}/\alpha_t )$ and $\hat{\x} := \mathrm{prox}_{\phi_{\frac{1}{\bar{\rho}}}}(\x)$ for any constant $\bar{\rho} > \rho$. 
Following Corollary~1, there exists $G, G_q >0 $ such that
$\mathbb{E}_t [\| \g(\w^{(Q)}_{t, G})\|^2] \leq G^2 + 2G_q^2$ and $\mathbb{E}_t\|\epsilon_{r, t} \|^2 \leq \alpha_t G_q^2$.
When $t \notin \mathcal{I}_E$, it holds that
\begin{align*}
    & \mathbb{E}_t [\phi_{\frac{1}{\bar{\rho}}}(\w^{(Q)}_{t+1, G})]\\
    & \leq \mathbb{E}_t [\mathcal{L}(\hat{\w}^{(Q)}_{t, G}) + \frac{\bar{\rho}}{2}\|\hat{\w }_{t, G} - \w^{(Q)}_{t+1, G} \|^2  ] \\
    & = \mathcal{L}(\hat{\w }^{(Q)}_{t, G}) + \mathbb{E}_t[ \frac{\bar{\rho}}{2}\|\hat{\w}^{(Q)}_{t, G} - \v_{t+1, G} \|^2 ] \\
    & = \mathcal{L}(\hat{\w }^{(Q)}_{t, G}) + \frac{\bar{\rho}}{2}\mathbb{E}_t [\| (\w^{(Q)}_{t, G} - \hat{\w }^{(Q)}_{t, G}) -  \alpha_{t} \g(\w^{(Q)}_{t, G}) \|^2 ] \\
    & \leq \mathcal{L}(\hat{\w}^{(Q)}_{t, G}) + \frac{\bar{\rho}}{2} \| \w^{(Q)}_{t, G} - \hat{\w }^{(Q)}_{t, G}\|^2 + \bar{\rho} \alpha_t \mathbb{E}_t\langle \hat{\w }^{(Q)}_{t, G} - \w^{(Q)}_{t, G}, \g(\w^{(Q)}_{t, G}) \rangle + \bar{\rho}(\alpha_t^2G^2+2\alpha_t G_q^2) \\
    & \leq \phi_{\frac{1}{\bar{\rho}}}(\w^{(Q)}_{t, G}) + \bar{\rho} \alpha_t \langle \hat{\w}^{(Q)}_{t, G} - \w^{(Q)}_{t, G}, v_t \rangle + \bar{\rho}(\alpha_t^2G^2+2\alpha_t G_q^2) \\
    & \leq  \phi_{\frac{1}{\bar{\rho}}}(\w^{(Q)}_{t, G}) + \bar{\rho}\alpha_t (\mathcal{L}(\hat{\w}^{(Q)}_{t, G} ) - \mathcal{L}(\w^{(Q)}_{t, G}) + \frac{\rho}{2} \|\w^{(Q)}_{t, G} - \hat{\w }^{(Q)}_{t, G} \|^2 ) + \bar{\rho}(\alpha_t^2G^2+2\alpha_t G_q^2),
\end{align*}
where the second inequality follows from the Young's inequality on the gradient variance, while the last inequality follows from the property of $\rho$-weakly convex functions. 

When $t \in \mathcal{I}_E$, we recall that $\v_{t, G} + \epsilon_{r, t} = \w_{t, G} $ and $\mathbb{E}_t[\epsilon_{r, t}] = 0$, leading to
\begin{align*}
    & \mathbb{E}_t [\phi_{\frac{1}{\bar{\rho}}}(\w^{(Q)}_{t+1, G})]\\
    & \leq \mathbb{E}_t [\mathcal{L}(\hat{\w}^{(Q)}_{t, G}) + \frac{\bar{\rho}}{2}\|\hat{\w }^{(Q)}_{t, G} - \w^{(Q)}_{t+1, G} \|^2  ] \\
    & = \mathcal{L}(\hat{\w }^{(Q)}_{t, G}) + \mathbb{E}_t[ \frac{\bar{\rho}}{2}\|\hat{\w}^{(Q)}_{t, G} - \v_{t+1, G} - \epsilon_{r, t} \|^2 ] \\
    & = \mathcal{L}(\hat{\w }^{(Q)}_{t, G}) + \frac{\bar{\rho}}{2}\mathbb{E}_t [\| (\w^{(Q)}_{t, G} - \hat{\w }^{(Q)}_{t, G}) -  \alpha_{t} \g(\w^{(Q)}_{t, G}) - \epsilon_{r} \|^2 ] \\
    & \leq \mathcal{L}(\hat{\w}^{(Q)}_{t, G}) + \frac{\bar{\rho}}{2} \| \w^{(Q)}_{t, G} - \hat{\w }^{(Q)}_{t, G}\|^2 + \bar{\rho} \alpha_t \mathbb{E}_t\langle \hat{\w }^{(Q)}_{t, G} - \w^{(Q)}_{t, G}, \g(\w^{(Q)}_{t, G}) \rangle + \bar{\rho}(\alpha_t^2G^2+3\alpha_t G_q^2) \\
    & \leq  \phi_{\frac{1}{\bar{\rho}}}(\w^{(Q)}_{t, G}) + \bar{\rho}\alpha_t (\mathcal{L}(\hat{\w}^{(Q)}_{t, G} ) - \mathcal{L}(\w^{(Q)}_{t, G}) + \frac{\rho}{2} \|\w^{(Q)}_{t, G} - \hat{\w }^{(Q)}_{t, G} \|^2 ) + \bar{\rho}(\alpha_t^2G^2+3\alpha_t G_q^2).
\end{align*}

Next, by using the fact that $\mathcal{L}(x) + \frac{\bar{\rho}}{2} \|x - x_t \|^2$ is $(\bar{\rho} - \rho)$-strongly convex we obtain 
\begin{align*}
    \mathcal{L}(\hat{\w}^{(Q)}_{t, G} ) - \mathcal{L}(\w^{(Q)}_{t, G}) + \frac{\rho}{2} \|\w^{(Q)}_{t, G} - \hat{\w }^{(Q)}_{t, G} \|^2 \leq \frac{\rho - \bar{\rho}}{\bar{\rho}^2} \|\nabla \phi_{\frac{1}{\bar{\rho}}} (\w^{(Q)}_{t, G}) \|^2.
\end{align*}
This further implies that
\begin{align*}
    \mathbb{E}_t [\phi_{\frac{1}{\bar{\rho}}}(\w^{(Q)}_{t+1, G})] & \leq  \phi_{\frac{1}{\bar{\rho}}}(\w^{(Q)}_{t, G}) - \alpha_t \frac{\bar{\rho} - \rho}{\bar{\rho}} \|\nabla \phi_{\frac{1}{\bar{\rho}}} (\w^{(Q)}_{t, G}) \|^2 + \bar{\rho}(\alpha_t^2G^2+3\alpha_t G_q^2).
\end{align*}
By telescoping and rearranging, we obtain
\begin{align*}
    & \frac{1}{\sum_{t=0}^T \alpha_{t}} \sum_{t=0}^T \alpha_{t} \mathbb{E} [ \|\nabla \phi_{\frac{1}{\bar{\rho}}} (\w^{(Q)}_{t, G}) \|^2 ] \\
    & \leq \frac{ \bar{\rho}}{\bar{\rho} - \rho} \frac{ \phi_{\frac{1}{\bar{\rho}}}(\w^{(Q)}_{0, G}) - \min \phi +  \bar{\rho}(G^2\sum_{t=0}^T \alpha^2_{t}+3G_q^2\sum_{t=0}^T \alpha_{t})}{\bar{\rho}\sum_{t=0}^T \alpha_{t}}.
\end{align*}

Recall that $\w_{t, G}$ is accessible only when $t \in \mathcal{I}_E$; hence, we consider only these time steps.
If we assume that at each communication round the client uses the same learning rate throughout $E$ local training epochs, we finally obtain that 
\begin{align*}
    & \frac{1}{\sum_{t=0}^T \alpha_{tE}} \sum_{t=0}^T \alpha_{tE} \mathbb{E} [ \|\nabla \phi_{\frac{1}{\bar{\rho}}} (\w^{(Q)}_{tE, G}) \|^2 ] \\
    & \leq \frac{E \bar{\rho}}{\bar{\rho} - \rho} \frac{ \phi_{\frac{1}{\bar{\rho}}}(\w^{(Q)}_{0, G}) - \min \phi +  \bar{\rho}(G^2\sum_{t=0}^T \alpha^2_{tE}+3G_q^2\sum_{t=0}^T \alpha_{tE})}{\bar{\rho}\sum_{t=0}^T \alpha_{tE} }.
\end{align*}
\end{proof}

While this result demonstrates convergence to a neighborhood of a stationary point, the specific SSL objective, $\mathcal{L}(\w) = \|\bar{X} - \w^T \w \|^2$, has property that all local minima are global minima \cite{jin2017escape}. The convergence to local minima via gradient descent using  bounded-variance gradient estimates has been studied in \cite{mertikopoulos2020almost} under more restrictive conditions including smoothness. It has been shown that for $\Gamma \geq \lambda_1(\bar{X})$, function $\mathcal{L}(\w)$ is $16\Gamma$-smooth within the region $\{\w | \|\w \|_2^2 \leq \Gamma \}$. Thus a suitable choice of $\Gamma$ and step size $\alpha_t$ guarantee that $\w_t$ remains within the region as long as $\w_{t-1}$ is within the region \cite{jin2017escape}. Therefore, Fed-QSSL approaches the neighborhood of the local minima of $\mathcal{L}(\cdot)$ and, ultimately, that of the global minima. 

    \subsection*{A.4. Proof of Theorem~2}
    {\textbf{Heterogeneous data construction. }}
The construction of heterogeneous data follows \cite{wang2022does}. Suppose there are $n = \Theta(d^{1/20})$ data sets/sources, one for each client, consisting of points that belong to $2n$ classes. For the data set at client $k$, the majority of points belong to classes $2k-1$ and $2k$, while a few remaining points come from other classes. We use $e_1, \dots, e_d$ to denote the standard unit basis of $\R^d$. A data sample from class $2k-1$ is generated as $x^{(2k-1)} = e_k - \sum_{i=1, i\neq k}^n q^{(2k-1, i)}\tau e_i + \mu \xi^{(2k-1)} $, where $q^{(2k-1, i)} \in \{0, 1 \} $ is uniformly sampled, $\xi^{(2k-1)} \sim \mathcal{N}(0, I)$, $\tau = d^{1/5}$ and $\mu = d^{-1/5}$. Likewise, a data sample from class $2k$ is generated as $x^{(2k)} = -e_k - \sum_{i=1, i\neq k}^n q^{(2k, i)}\tau e_i + \mu \xi^{(2k-1)} $. The numbers of data points in classes $2k-1$ and $2k$ are the same and of order $poly(d)$. As for the other classes, a data sample from class $2i-1$ is generated as $x^{(2i-1)} = e_i + \mu \xi^{(2i-1)}$ when $i \neq k$ and there are no data samples in class $2i$. The numbers of data points in class $2k-1$ and class $2k$ are the same and of order $\O(d^{\beta})$ for some $\beta \in (0, 1)$ such that $\O(nd^{\beta}) \leq \O(d^{1/5})$, implying that the total number of data samples in infrequent classes is much smaller than the number of data samples in the frequent classes (i.e., in classes $2k-1$ and $2k$). For simplicity, we assume that all data sets are of the same cardinality, i.e., $\frac{|D_k|}{|D|} = \frac{1}{n}$ for all $k \in [n]$.

The following theorem quantifies the robustness property.

\newtheorem*{theorem2}{Theorem 2}
\begin{theorem2}
Suppose assumptions of Theorem~1 hold and $n = \Theta(d^{1/20})$. In a $2n$-way classification task, when in local training the update is $\epsilon$ away from the optimal solution $\w_k^*$, the representation vector learned by client $k$ with high probability satisfies $ \frac{d^{2/5} - \O(d^{-2/5}) + 2e_j^T (\w_k^*)^T \epsilon e_j + (e_j^T \epsilon)^2}{d^{2/5} + \O(d^{-2/5}) + \|2(\w_k^*)^T \epsilon + \epsilon^T \epsilon  \|} \leq r_{i}^k \leq 1$ for $i \in [n]\backslash \{k \}$. As for the global objective, when the update is $\epsilon_1$ away from the optimal solution $\w^*$, the learned representation vector $\bar{r}$ with high probability satisfies 

$\frac{d^{2/5} - \Theta(d^{7/20}) + \O(d^{-1/20}) - \O(d^{2/5}) + 2e_j^T (\w^*)^T \epsilon_1 e_j + (e_j^T \epsilon_1)^2 }{d^{2/5} - \Theta(d^{7/20}) + \O(d^{-1/20}) + \|2(\w^*)^T \epsilon_1 + \epsilon_1^T \epsilon_1  \|} \leq \bar{r}_i \leq 1$ for all $i \in [n]$. 
\end{theorem2}

{\textbf{Technical challenges.}}
While the robustness of $\w^*$ was analyzed in \cite{wang2022does}, it is unclear how to extended it to quantifying robustness when the update obtained by the proposed algorithm approaches the optimal solution. To facilitate such analysis, we initiate it at the step the obtained update is $\epsilon_1$ distance away from the optimal $\w^*$; this maps the problem to that of analyzing perturbed representation vectors and data covariance matrix. We conduct such analysis by borrowing ideas from matrix perturbation theory. 

\begin{proof}
For the local data set/source at client $k$, 
\[
X_k = \mathbb{E}_{x \sim D_k}(xx^T) = \frac{1}{|D_k|}\sum_{i=1}^{|D_k|} x_{k, i}x_{k, i}^T
\]
is the empirical covariance matrix. Its expectation is
\begin{small}
\begin{equation}
\begin{aligned}
    \mathbb{E}[X_k] & = \mathrm{diag}(\tau^2 + O(d^{-2/5}), \cdots, 1+O(d^{-2/5}), \cdots, \tau^2 + O(d^{-2/5}),  O(d^{-2/5}) \\
    & \quad \cdots,  O(d^{-2/5}) ) \\
    & = \mathrm{diag}( d^{2/5}+O(d^{-2/5}), \cdots, 1+O(d^{-2/5}), \cdots, d^{2/5}+O(d^{-2/5}), O(d^{-2/5}), \\ 
    & \quad \cdots, O(d^{-2/5})  ), \label{eq:r1}
\end{aligned}
\end{equation}
\end{small}
where the first $n$ diagonal entries in the first equality are $ \tau^2+ O(d^{-2/5})$, except the $k$-th entry which is $ 1+ O(d^{-2/5})$; the remaining $d-n$ diagonal entries are $ O(d^{-2/5})$.

Using matrix concentration bounds \cite{vershynin2018high} and Weyl's inequality, we have that with high probability 
\begin{align}\label{eq:r2}
    |\lambda_{k, i} - \lambda_i(\mathbb{E}X_k)| \leq \|X_k - \mathbb{E}X_k \|_2 \leq O(d^{-2/5}),
\end{align}
where $\lambda_{k, i}$ represents the $i$-th largest eigenvalue of $X_k$.

Since $|D_k|$ is of the order $poly(d)$, one can leverage Lemma~E.1 in \cite{liu2021self} to obtain that $\sum_{i=1}^{|D_k|} |\langle \xi_{k, i}, e_k \rangle| /|D_k| \leq O(d^{1/10})$. For $\mu = d^{-1/5}$,
\begin{align*}
    e_j^T X_k e_j  = \mathbb{E}[(e_j^Tx)^2] \geq [\mathbb{E}(e_j^T x)]^2 \geq (\frac{1}{3}\tau - \mu \sum_{i=1}^{|D_k|} \frac{1}{|D_k|}|e_j^T \xi_{k, i}| ) = \Omega(\tau^2) = \Omega(d^{2/5})
\end{align*}
with probability at least $1-\frac{1}{2}e^{-d^{1/10}}$. The second inequality follows from the fact that for at least $1/3$ data points it holds that either $q^{2k-1, j}$ or $q^{2k, j}$ is $1$. From \eqref{eq:r1} and \eqref{eq:r2}, one can further obtain that
\begin{equation}
\begin{aligned}
    e_j^T X_k e_j & = e_j^T \mathbb{E}[X_k] e_j + e_j^T (X_k - \mathbb{E}(X_k)) e_j  \\
    & \geq d^{2/5} + \O(d^{-2/5}) - \|X_k - \mathbb{E}(X_k) \| \geq d^{2/5} - \O(d^{-2/5}), \label{eq:r3}
\end{aligned}
\end{equation}
and
\begin{equation}
    \begin{aligned}
        \lambda_{k, 1} \leq \lambda_1(\mathbb{E}(X_k)) + \O(d^{-2/5}) = d^{2/5} + \O(d^{-2/5}).
    \end{aligned}
\end{equation}

Recall that the local optimization problem can equivalently be written as the minimization of a local objective function $\mathcal{L}_k = \|X_k - \w^T \w \|^2$. By Eckart-Young-Mirsky theorem, minimizing $\mathcal{L}_k$ leads to solution $\w_k^* \in \R^{m \times d}$ whose rows are eigenvectors of the first $m$ eigenvalues of $X_k$. Let $\{v_{k, 1}, \cdots, v_{k, d} \}$ be the set of $d$ orthonormal eigenvectors of $X_k$, $X_k = \sum_{i=1}^d \lambda_{k, i} v_{k, i}v_{k, i}^T$, and $(\w^*_k)^T(\w^*_k) = \sum_{i=1}^m \lambda_{k, i} v_{k, i}v_{k, i}^T $. We readily obtain that 
\begin{align}\label{eq:r4}
    \lambda_{k, 1} \sum_{i=1}^d (e_j^T v_{k, i})^2 \geq e_j^T X_k e_j = \sum_{i=1}^d \lambda_{k, i} (e_j^T v_{k, i})^2
\end{align}
with probability of at least $1-\frac{1}{2}e^{-d^{1/10}}$. Suppose the update is within $\epsilon$ distance from the optimal solution, i.e., $\w_{k, t} = \w_k^* + \epsilon$. Let $\w_{k, t}^T \w_{k, t} = \sum_{i=1}^m \lambda'_{k, i} (v'_{k, i})^T v'_{k, i}$. For $i > m$, we define $v'_{k, i} = v_{k, i}$ and $\lambda'_{k, i} = \lambda_{k, i}$. Invoking result from matrix perturbation theory,
\begin{align}\label{eq:matrix}
    \max_{1\leq i \leq d} |\lambda_t(A+E) - \lambda_t(A)| \leq \|E \|_2 \leq \|E \|,
\end{align}
which implies that $ |\lambda_{k, i} - \lambda'_{k, i} |$ is upper bounded by $\|2(\w_k^*)^T \epsilon + \epsilon^T \epsilon\|$ for any $i \in [n]$. From \eqref{eq:r3}-\eqref{eq:r4} and $\w_{k, t} = \w_k^* + \epsilon $, it follows that 
\begin{align*}
    \sum_{i=1}^d (e_j^T v'_{k, i})^2 & \geq \frac{e_j^T X_k e_j + 2e_j^T (\w_k^*)^T \epsilon e_j + (e_j^T \epsilon)^2 }{\lambda'_{k, 1}} \\
    & \geq \frac{e_j^T X_k e_j + 2e_j^T (\w_k^*)^T \epsilon e_j + (e_j^T \epsilon)^2}{\lambda_{k, 1} + \|2(\w_k^*)^T \epsilon + \epsilon^T \epsilon \|} \\
    & \geq \frac{d^{2/5} - \O(d^{-2/5}) + 2e_j^T (\w_k^*)^T \epsilon e_j + (e_j^T \epsilon)^2}{d^{2/5} + \O(d^{-2/5}) + \|2(\w_k^*)^T \epsilon + \epsilon^T \epsilon \|}.
\end{align*}
Let $r^k = [r_1^k, \cdots, r_d^k ]^T $ be the representation vector learned at client $k$. Then $\forall i \in [n]/\{k \}$,
\begin{align*}
    \frac{d^{2/5} - \O(d^{-2/5}) + 2e_j^T (\w_k^*)^T \epsilon e_j + (e_j^T \epsilon)^2}{d^{2/5} + \O(d^{-2/5}) + \|2(\w_k^*)^T \epsilon + \epsilon^T \epsilon  \|} \leq r_{i}^k.
\end{align*}


Next we move to the global objective, $ \mathcal{L} = \|\bar{X} - \w^T \w \|^2$, where $\bar{X} = \mathbb{E}_{x \sim D}(xx^T) = \sum_k \frac{|D_k|}{|D|}X_k = \frac{1}{|D|}\sum_{i=1}^{|D|}x_i x_i^T$ is the empirical global data covariance matrix. Following the linearity of expectation and the fact that 
\begin{align*}
    \frac{(n-1)d^{2/5}+1}{n} & = (1 - \Theta(d^{-1/20}))d^{2/5} + \O(d^{-1/20}) \\
    & = d^{2/5} - \Theta(d^{7/20}) + \O(d^{-1/20}),
\end{align*}
we obtain the expectation of the covariance matrix as
\begin{align*}
    \mathbb{E}[{\bar{X}}] & = \mathrm{diag}( d^{2/5} - \Theta(d^{7/20}) + \O(d^{-1/20}), \cdots, d^{2/5} - \Theta(d^{7/20}) + \O(d^{-1/20}), \\
    & \quad \O(d^{-2/5}), \cdots, \O(d^{-2/5}) )
\end{align*}
where the first $n$ entries are of the same order while the remaining $d-n$ entries are $\O(d^{-2/5})$. Let $\{v_{1}, \cdots, v_{d} \}$ be the set of $d$ orthonormal eigenvectors of $\bar{X} = \sum_{i=1}^d \lambda_{i} v_{i}v_{i}^T$, and $(\w^*)^T(\w^*) = \sum_{i=1}^m \lambda_{i} v_{i}v_{i}^T $. 

Follow similar techniques to analyze local objectives leads to
\begin{align*}
    e_j^T \bar{X} e_j & \geq d^{2/5} - \Theta(d^{7/20}) + \O(d^{-1/20}) - \O(d^{2/5}) \\
    \lambda_1(\bar{X}) & \leq \lambda_1(\mathbb{E}(\bar{X})) + \O(d^{2/5}) = d^{2/5} - \Theta(d^{7/20}) + \O(d^{-1/20}).
\end{align*}

Next, we consider perturbation $\w_{t} = \w^* + \epsilon_1 $ of the global optimal solution $\w^*$. Let $\w_{t}^T \w_{t} = \sum_{i=1}^m \lambda'_{i} (\bar{v}'_{i})^T \bar{v}'_{i}$. For $i > m$, denote $\bar{v}'_{i} = \bar{v}_{i}$ and $\lambda'_{i} = \lambda_{i}$. Then
\begin{align*}
    \sum_{i=1}^d (e_j^T \bar{v'}_{i})^2 & \geq \frac{e_j^T \bar{X} e_j + 2e_j^T (\w^*)^T \epsilon_1 e_j + (e_j^T \epsilon_1)^2 }{\lambda'_{1}} \\
    & \geq \frac{e_j^T \bar{X} e_j + 2e_j^T (\w^*)^T \epsilon_1 e_j + (e_j^T \epsilon_1)^2 }{\lambda_{1} + \|2(\w^*)^T \epsilon_1 + \epsilon_1^T \epsilon_1 \|} \\
    & \geq \frac{d^{2/5} - \Theta(d^{7/20}) + \O(d^{-1/20}) - \O(d^{2/5}) + 2e_j^T (\w^*)^T \epsilon_1 e_j + (e_j^T \epsilon_1)^2 }{d^{2/5} - \Theta(d^{7/20}) + \O(d^{-1/20}) + \|2(\w^*)^T \epsilon_1 + \epsilon_1^T \epsilon_1  \|}
\end{align*}
where the second inequality is due to 
the matrix perturbation result \eqref{eq:matrix} applied to the global data covariance matrix.

We then let $\bar{r} = [\bar{r}_1, \cdots, \bar{r}_d ]^T$ be the representation vector learned from the global objective and $\forall i \in [n]$,
\begin{align*}
    \frac{d^{2/5} - \Theta(d^{7/20}) + \O(d^{-1/20}) - \O(d^{2/5}) + 2e_j^T (\w^*)^T \epsilon_1 e_j + (e_j^T \epsilon_1)^2 }{d^{2/5} - \Theta(d^{7/20}) + \O(d^{-1/20}) + \|2(\w^*)^T \epsilon_1 + \epsilon_1^T \epsilon_1  \|} \leq \bar{r}_i 
\end{align*}

\end{proof}

\begin{algorithm}[tb]
\renewcommand{\thealgorithm}{2}
\caption{Fed-QSSL local training: client $k$ trains on $L$-layer network with $s_k$-bit weights and activations using $s_k+2$-bit gradients.}
\label{alg:algorithm3}
\textbf{Require}: a minibatch of inputs $a_0$ , previous weights $\w^{(Q)}_{k, t}$ in $s_k$-bit, learning rate $\alpha_t$, activation function $h$ \\
\begin{algorithmic}[1] 
\STATE $\{$ Forward pass $\}$
\FOR{$l = 1$ to $L$}
\STATE $\Tilde{a}_l \leftarrow $ forward($a_{l-1}, (\w^{(Q)}_{k, t})_{l}$)
\STATE $a_{l} \leftarrow h(\Tilde{a}_l )$
\STATE Compute $a^{(Q)}_{l}$ through $\tanh$-based quantizer
\ENDFOR
\STATE $\{$ Back propagation $\}$
\STATE Compute the gradient $g_{a_L}$ at the last layer and then compress to $g^{(Q)}_{a_L}$
\FOR{$l = L$ to $1$}
\STATE 
$g_{a_{l-1}} \leftarrow  $ inputs backpropagation$(g^{(Q)}_{a_l}, (\w^{(Q)}_{k, t})_{l} ) $
\STATE 
$(g_{\w_{k, t}})_{l} \leftarrow $ weights backpropagation$(g^{(Q)}_{a_l}, a^{(Q)}_{l-1}) $
\STATE Compute $g^{(Q)}_{a_{l-1}} $ via quantile quantization
\STATE Compute $(g^{(Q)}_{k, t})_{l}$ via quantile quantization
\ENDFOR
\STATE $\{$ Weight update $\}$
\STATE $ \w^{(Q)}_{k, t+1} = Q(\w^{(Q)}_{k, t} - \alpha_t \g^{(Q)}_{k, t}) $
\end{algorithmic}
\end{algorithm}

    \subsection*{A.5. Local Training Implementation Details}
    Details of local quantized training (steps 5-6 in Fed-QSSL in the main manuscript) are specified in Algorithm \ref{alg:algorithm3}. In particular, in the forward pass we go from the first layer to the last one, obtain quantized activations along the way; in the backward propagation, we go from the last layer back to the first layer, computing and quantizing gradients in the process.

	\clearpage
	\bibliography{bib}
	\bibliographystyle{acm}
    
\end{document}